\documentclass{cmmse2014}
\usepackage{amssymb}
\usepackage[pdftex]{graphicx}
\usepackage{epstopdf}
\usepackage{color}
\usepackage{amsmath}
\usepackage{amsfonts}
\usepackage{amssymb}
\usepackage{caption}
\usepackage{graphicx}
\usepackage{float}
\usepackage{multirow}

\usepackage{filecontents}

\begin{document}


\newcommand{\Eqref}[1]{(\ref{#1})}
\newcommand\thefont{\expandafter\string\the\font}

\newcommand*{\underuparrow}[1]{\ensuremath{\underset{\uparrow}{#1}}}
\captionsetup{width=0.8\textwidth}


\markboth
  {} 
  {}


\title{Assessing the effect of advertising expenditures upon sales: a Bayesian structural time series model}



\author{V\'ictor Gallego}{victor.gallego@icmat.es}{1}

\author{Pablo Su\'arez-Garc\'{\i}a}{pasuarez@ucm.es}{2}

\author{Pablo Angulo}{pablo.angulo@upm.es}{3}

\author{David G\'omez-Ullate}{david.gomez-ullate@icmat.es}{{1,2,4}}


\affiliation{1}{Institute of Mathematical Sciences}
  {ICMAT-CSIC}

\affiliation{2}{Facultad de Ciencias F\'isicas}{Universidad Complutense de Madrid}

\affiliation{3}{ETSIN}{Universidad Polit\'ecnica de Madrid}
\affiliation{4}{Department of Computer Science, School of Engineering}{Universidad de C\'adiz}

\begin{abstract}
We propose a robust implementation of the Nerlove--Arrow model using a Bayesian structural time series model to explain the relationship between advertising expenditures of a country-wide fast-food franchise network with its weekly sales. Thanks to the flexibility and modularity of the model, it is well suited to generalization to other markets or situations. Its Bayesian nature facilitates incorporating \emph{a priori} information reflecting the manager's views, which can be updated with relevant data. This aspect of the model will be used to support the decision of the manager on the budget scheduling of the advertising firm across time and channels.


\end{abstract}

%
%

\section{Introduction}

It is widely acknowledged that a firm's expenditure  on advertising has a positive effect on sales \cite{assmus1984advertising, tellis2007advertising, luo2012does, wiesel2011practice}. However, the exact relationship between them remains a moot point, see \emph{e.g.} \cite{tellis2009generalizations} for a broad survey. Since Dorfman and Steiner's \cite{dorfman1954optimal} seminal work\,---one of the first formal theories of optimal monopoly advertising---\,several models have been proposed to pinpoint this relationship, although consensus on the best approach has not been reached yet. Two diverging model-building schools seem to dominate the marketing literature \cite{little1979aggregate}: \emph{a priori} models that rely heavily on intuition and are derived from general principles, although usually with a practical implementation on mind (\cite{nerlove1962optimal}, or \cite{vidale1957operations} and \cite{little1975brandaid} \emph{inter alia}) and  \emph{statistical} or \emph{econometric} models, which usually start from a specific dataset to be modelled (see \cite{assmus1984advertising} for a review). In this work we will mostly rely on the first type of models, \emph{viz.} that of Nerlove-Arrow \cite{nerlove1962optimal}, which extends  Dorfman and Steiner to a dynamic setting \cite{bagwell2007economic} and adapts seamlessly to the \emph{state-space} or \emph{structural} time series approach.

Bayesian structural time series models \cite{scott2014predicting}, in turn, have positioned themselves in the past few years as very effective tools  not only for analysing marketing time-series, but also to throw light into more uncertain terrains like  causal impacts, incorporating \emph{a priori} information into the model, accommodating multiple sources of variations or supporting variable selection. Although the origins of this formalism can be traced back to the 1950's in engineering problems of filtering, smoothing and forecasting, first with Wiener \cite{wiener1949extrapolation} and specially with Kalman \cite{kalman1960new}, these problems can also be understood from the perspective of estimation in which a vector valued time series $\{ X_0, X_1, X_2, \ldots\}$ that we wish to estimate (the \emph{latent} or \emph{hidden} states) is observed through a series of noisy measurements $\{ Y_0, Y_1, Y_2, \ldots\}$. This  means that, in the Bayesian sense, we want to compute the joint posterior distribution of all the states given all the measurements \cite{sarkka2013bayesian}. The ever-growing computing power and release of several programming libraries  in the last few years like \cite{petris2010r, scott2016bsts} have in part alleviated the difficulties in the mathematical underpinning and computer implementation  that this formalism suffers, making these methods broadly known and used. This family of models have been used successfully to model financial time series data \cite{doi:10.1002/asmb.428}, infer causal impact of marketing campaigns \cite{brodersen2015inferring}, select variables and nowcast consumer sentiment  \cite{scott2015bayesian}, or for predicting other economic time series models like unemployment \cite{scott2014predicting}.

In this paper we use the formalism of Bayesian structural time-series models to formulate a robust model that links advertising expenditures with weekly sales. Due to the flexibility and modularity of the model, it will be well suited to generalization to various markets or situations. Its Bayesian nature also adapts smoothly to the issue of introducing outside or \emph{a priori} information, which can be updated to the posterior distribution of the estimated parameters. The formulation of the model allows for non-gaussian innovations of the process, which will take care of the heavy-tailedness of the distribution of sales increments. We also discuss how the forecasts produced by this model can help the manager into allocating the advertising budget. The decision space is reduced to a one dimensional curve of Pareto optimal strategies for the two moments of the forecast distribution: expected return and variance.

The paper is organized as follows: after a brief review of the most usual marketing models and the formalism of structural time series in section 2, we define the model to be used to fit the data. The experimental setup will be detailed in section 3. Section 4 provides a discussion in which alternative models will be compared and also possible uses of this model in the industry. A brief summary and ideas for further research are detailed in section 5.

\section{Theoretical background and model definition}

\subsection{The Nerlove-Arrow model}

Numerous formulations of aggregate advertising response models exist in the marketing literature, e.g. \cite{little1979aggregate}. The model of Nerlove and Arrow \cite{nerlove1962optimal} extends the Dorfman-Steiner model to cover the situation in which present advertising expenditures affect future demand for products; it is parsimonious and is considered as a standard in the quantitative marketing community. We use it as our starting point.

In this model, advertising expenditures are considered similar in many ways to investments in durable plant and equipment, in the sense that they affect the present and future character of output and, hence, the present and future net revenue of the investing firm. The idea is to define an ``advertising stock'' called \emph{goodwill}  $A(t)$ which seemingly summarizes the effects of current and past advertising expenditures over demand. Then, the following dynamics is defined for the goodwill

\begin{equation}\label{eq:NA}
\frac{dA}{dt} = qu(t) - \delta A(t),
\end{equation}
where $u(t)$ is the advertising spending rate (e.g., euros or gross rating points per week), $q$ is a parameter that reflects the advertising quality (an effectiveness coefficient) and $\delta$ is a decay or forgetting rate. The \emph{goodwill} then increases linearly with the advertisement expenditure but decreases also linearly due to forgetting. 

Several extensions and modifications have been proposed to this simple model: it can include a limit for potential costumers \cite{vidale1957operations}, a non-linear response function to advertise expenditures \cite{little1975brandaid}, wear-in and wear-out effects of advertising \cite{naik1998planning},  interactions between different advertising channels \cite{bass2007wearout}, among other. Still, for most of the tasks, the  Nerlove-Arrow model remains as a simple and solid starting point. 

\subsection{Bayesian structural time series models}

\emph{Structural time series models} or \emph{state-space models} provide a general formulation that  allows a unified treatment of virtually any linear time series model through the general algorithms of the Kalman filter and the associated smoother. Several handbooks \cite{durbin2012time, petris2009dynamic, sarkka2013bayesian, west1998bayesian} discuss this topic in depth, so  we will not develop the corresponding theory. We will however present the most salient features that concern our modelling problem. For further details, the reader may check the aforementioned handbooks. 

The state-space formulation of a time series consists of two different equations: the \emph{state} or \emph{evolution equation} which determines the dynamics of the state of the system as a first-order Markov process ---\,usually parametrized through \emph{state} variables\,--- and an \emph{observation} or \emph{measurement  equation} which links the latent state with the observed state. Both equations are also affected by noise. Denoting by $\mathbf{\theta}_t$ the $m\times1$ \emph{state vector} describing the inner state of the system, by $G_t$ the $m\times m$ matrix that generates the dynamics, by $H_t$ a $m\times g$ matrix and by $ \epsilon_t $ a $g\times 1$ vector of serially uncorrelated disturbances with mean zero and covariance matrix $W_t$, the system would evolve according to the equation

\begin{equation}\label{eq:st}
\theta_{t} = G_t \theta_{t-1} + H_t \epsilon_t \qquad \epsilon_t \sim \mathcal{N}(0, W_t).
\end{equation}
The states ($\theta_t$) are not generally observable, but are linked to the \emph{observation variables} $Y_t$ through the \emph{observation equation}:
\begin{equation}\label{eq:obs}
Y_t = F_t \theta_t + \epsilon'_t \qquad \epsilon'_t \sim \mathcal{N}(0, V_t),
\end{equation}
where, $Y_t \in \mathbb{R}$ is the observed value at timestep $t$, $F_t$ is the $1\times m$ matrix that links the inner state to the observable, and $V_t \in \mathbb{R}^+$ is the variance of $\epsilon'_t $, the random disturbances of the observations.



The specification of the state-space system is completed by assuming that the initial state vector $\theta_0$ has mean $\mu_0$ and a covariance matrix $\Sigma_0$ and it is uncorrelated with the noise. The problem then consists of \emph{estimating the sequence of states $\{\theta_1, \theta_2, \ldots\}$ for a given series of observations $\{y_1, y_2, \ldots\}$} and whichever other structural parameters of the transition and observation matrices. State estimation is readily performed via the \emph{Kalman filter};  different alternatives however arise  when structural parameters are unknown. In the classical setting, these are estimated using maximum likelihood. In the Bayesian approach, the probability distribution about the unknown parameters is updated via Bayes Theorem. If exact computation through conjugate priors is not possible, the probability distributions before each measurement are updated by approximate procedures such as Markov chain Monte Carlo (MCMC) \cite{scott2014predicting}. 

The Bayesian approach offers several advantages compared to classical methods. For instance, it is natural to incorporate external information through the prior distributions. In particular, this will be materialized in Section \ref{sec:s_s} where expert information can be incorporated through the spike and slab prior. Another useful advantage is that, due to the Bayesian nature of the model, it is straightforward to obtain predictive intervals through the predictive distribution (see Section \ref{s:MCMC}).

\subsection{Model specification}\label{sec:model}

\subsubsection{State-space formulation}

The continuous-time Nerlove-Arrow model must be first cast in discrete time so as to formulate our model in state-space. From equation (\ref{eq:NA}), we get:

$$
A_t =  (1-\delta) A_{t-1} + q u_{t-1} + \epsilon_t 
$$
where $A_t$ is the \emph{goodwill} stock, $u_t$ is the advertising spending rate, $q$ is the effectiveness coefficient and the random disturbance $\epsilon_t$  captures the net effects of the variables that affect the goodwill  but cannot be modelled explicitly.  This discrete counterpart of Nerlove-Arrow is a distributed-lag structure with geometrically declining weights, i.e., a \emph{Koyck model} \cite{clarke1976econometric, koyck1954distributed}. Since in our setting the model includes the effect of $k$ different channels in the goodwill, we shall modify the previous equations according to:

$$
A_t =  (1-\delta) A_{t-1} + \sum_{i=1}^k q_i u_{i(t-1)} + \epsilon_t 
$$

Now, following the notation from (\ref{eq:st}) and (\ref{eq:obs}), the discrete-time Nerlove-Arrow model in state-space form will read as follows:
\begin{description}

\item[Evolution equation:]

\begin{equation} \label{eq:NA_st}
\theta_{t} = G_t \theta_{t-1} + H_t \epsilon_t \qquad \epsilon_t \sim \mathcal{N}(0, W_t),   
\end{equation}
\noindent where
\begin{equation*} 
\theta_{t}  = \begin{bmatrix} A_t  \\ q_1 \\  \vdots \\q_k \end{bmatrix}
, \quad
G_t =  \begin{bmatrix}
   (1- \delta) & u_{1(t-1)} &  \ldots & u_{k(t-1)} \\
   0 & 1 &   \ldots & 0 \\
   \vdots  &   \vdots & \ddots & \vdots \\
   0 & 0 &  \ldots & 1 \\
   \end{bmatrix}
, \quad
 H_t =    \begin{bmatrix} 1  \\ 0 \\ \vdots \\ 0 \end{bmatrix}
\end{equation*}

Note that the $q_i$ are constant over time and that the matrix $G_t$ depends on the known inversion levels at time $t-1$ and on an unknown parameter ($\delta$) to be estimated from the data.

 \item[Observation equation:]
 
 \begin{equation} \label{eq:NA_obs}
 Y_t = F_t \theta_t + \epsilon'_t \qquad \epsilon'_t \sim \mathcal{N}(0, V_t)  
 \end{equation}

 where $Y_t$ are the observed sales at time $t$ and $F_t = \begin{bmatrix} 1,  & 0 , & \ldots, & 0 \end{bmatrix}$.
\end{description}

\subsection{Modularity and additional structure}

The above model is very flexible in the sense that it can be defined \emph{modularly},  in as much as the different hidden states evolve independently of the others (\emph{i.e.} the evolution matrix can be cast in block-diagonal form). This greatly simplifies their implementation and allows for simple building-blocks with characteristic behavior. Typical blocks specify trend and seasonal components ---\,which can be helpful to discover additional patterns in the time series--- or explanatory variables that can be added to further reduce the uncertainty in the model and bridge the gap between time series and regression models. Via the \emph{superposition principle} \cite[Chapter 3]{petris2009dynamic} we could include additional blocks in our model:

$$
Y_t = Y_{NA, t} + Y_{R, t} + Y_{T, t} + Y_{S, t}
$$
where $Y_{NA, t}$ corresponds to the discretized Nerlove-Arrow equation, defined in (\ref{eq:NA_st}) and (\ref{eq:NA_obs}); $Y_{R, t}$ is a regression component; containing the effects of  external explanatory variables $X_t$; $Y_{T, t}$ is a trend component or a simpler local level component; and $Y_{S, t}$ is a seasonal component.

\subsubsection{Regression components. Spike and slab variable selection}\label{sec:s_s}

To take into account the effects of external explanatory variables, a static regression component can be easily incorporated into the model through

$$
Y_{R,t} = X_t \beta + \epsilon_t,
$$
where the state $\beta$ is constant over time to favor parsimony.

A spike and slab prior \cite{mitchell1988bayesian} is used for the static regression component since it can incorporate \emph{prior information} and also facilitate variable selection. This is specially useful for models with large number of regressors, a typical setting encountered in business scenarios.
Let $\gamma$ denote a binary vector that indicates whether the regressors are included in the regression. Specifically, $\gamma_i = 1$ if and only if $\beta_i \neq 0$. The subset of $\beta$ for which $\gamma_i = 1$ will be denoted $\beta_{\gamma}$. Let $\sigma^2_{\epsilon}$ be the residual variance from the regression part. The spike and slab prior \cite{george1997approaches} can be expressed as
$$
p(\beta, \gamma, \sigma^2_{\epsilon}) = p(\beta_{\gamma} | \gamma, \sigma^2_{\epsilon})p(\sigma^2_{\epsilon} | \gamma)p(\gamma).
$$
A usual choice for the $\gamma$ prior is a product of Bernoulli distributions:

$$
\gamma \sim \Pi_i \pi_i^{\gamma_i}(1-\pi_i)^{1-\gamma_i}.
$$
The manager of the firm may elicit these $\pi_i$ in various ways. A reasonable choice when detailed prior information is unavailable is to set all $\pi_i = \pi$. Then, we may specify an expected number of non-zero coefficients by setting $\pi = k/p$, where $p$ is the total number of regressors. Another possibility is to set $\pi_i = 1$ if the manager believes that the $i-$th regressor is crucial for the model. 

\subsection{Model estimation and forecasting}\label{s:MCMC}

Model parameters can be estimated using Markov Chain Monte Carlo simulation, as described in Chapter 4 of \cite{petris2009dynamic} or \cite{scott2014predicting}. We follow the same scheme.

Let $\Theta$ be the set of model parameters other than $\beta$ and $\sigma^2_{\epsilon}$. The posterior distribution can be simulated with the following Gibbs sampler: 

\begin{enumerate}
\item Simulate $\theta \sim p(\theta | y, \Theta, \beta, \sigma^2_{\epsilon})$.
\item Simulate $\Theta \sim p(\Theta | y, \theta, \beta, \sigma^2_{\epsilon})$.
\item Simulate $\beta, \sigma^2_{\epsilon} \sim p(\beta, \sigma^2_{\epsilon} | y, \theta, \Theta)$.
\end{enumerate}
Repeatedly iterating the above steps gives a sequence of draws $\rho^{(1)}, \rho^{(2)}, \ldots, \rho^{(K)}$ $\sim$ $p(\Theta, \beta, \sigma^2_{\epsilon}, \theta)$. In our experiments, we set $K = 4000$ and discard the first 2000 draws to avoid burn-in issues. \\

In order to sample from the predictive distribution, we follow the usual Bayesian approach summarized by the following predictive equation, in which $y_{1:t}$ denotes the sequence of observed values, and $\bar{y}$ denotes the set of values to the forecast

\[
p(\bar{y} | y_{1:t}) = \int p(\bar{y}| \rho)p(\rho | y_{1:t}) d\rho.
\]
Thus, it is sufficient to sample from $p(\bar{y}| \rho^{(i)})$, which can be achieved by iterating equations (\ref{eq:obs}) and (\ref{eq:st}). With these predictive samples $\bar{y}^{(i)}$ we can compute statistics of interest regarding the predictive distribution   $p(\bar{y} | y_{1:t})$ such as the mean or variance (MC estimates of $E[ \bar{y} | y_{1:t}]$ and $Var[\bar{y} | y_{1:t}]$, respectively) or quantiles of interest.\\

\subsubsection{Robustness}

We can replace the assumption of Gaussian errors with student-$t$ errors in the observation equation, thus leading to the model
$$ 
Y_{t} = F_t \theta_t + \epsilon'_t \qquad \epsilon'_t \sim \mathcal{T}_\nu(0, \tau^2).
$$
Typically, in these settings we set $\nu > 1$ to make the variance finite, and this variance parameter can be estimated from data using Empirical Bayes methods, for instance.
In this manner, we allow the model to predict occasional larger deviations, which is reasonable in the context of forecasting sales. For instance, a special event not taken into account through the predictor variables may lead to an increase in the sales for that week.


\section{Case Study. Data and parameter estimation}

\subsection{Data analysis}

\begin{figure}[h]
\centering
\includegraphics[scale=0.6]{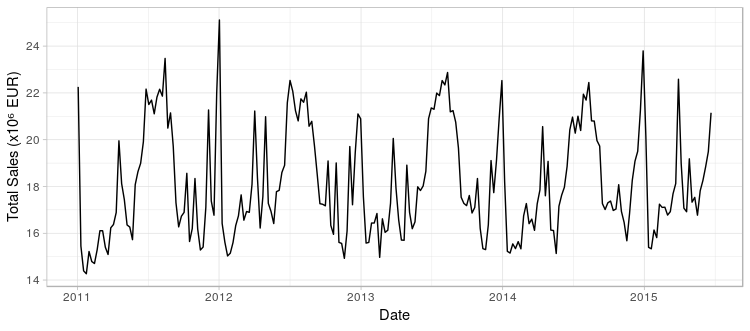}
\caption{Total weekly sales. Jan-2011 to Jun-2015}\label{fig:sales}
\end{figure}

\begin{figure}[h]
\centering
\includegraphics[scale=0.5]{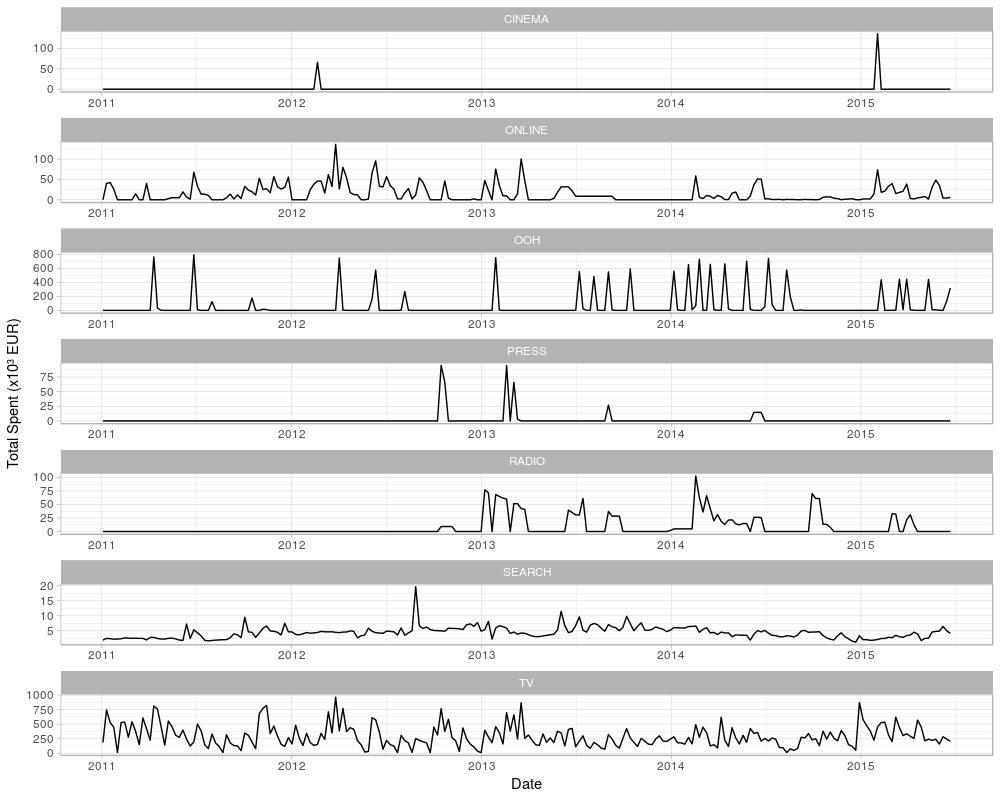}
\caption{Advertising expenditure. Jan-2011 through Jul-2015. \\ Note the different scales on the y-axis.}\label{fig:canales}
\end{figure}

The time series analyzed in this case study contains the total weekly sales of a country-wide franchise of fast food restaurants, Figure \ref{fig:sales}, covering the period January 2011 - June 2015, thus comprising 234 observations. The total weekly sales is in fact the aggregated sum from the individual sales of the whole country network of 426 franchises, also allowing a fine-grained study down to the store level, although we shall not carry it out in this paper. Along with the sales figures, the series includes the investment levels $\{u_{it}\}$  in advertising during this period  for seven different channels \emph{viz.} \texttt{OOH} (Out-of-home, \emph{i.e.} billboards), \texttt{Radio}, \texttt{TV}, \texttt{Online}, \texttt{Search}, \texttt{Press} and \texttt{Cinema}, Figure \ref{fig:canales}, $i=1,\ldots,7$.

From such graphs we observe that:
\begin{itemize}
\item In Figure \ref{fig:sales}, the series has a characteristic seasonal pattern that shows peaks in sales coinciding with Christmas, Easter and summer vacations. 
\item In Figure \ref{fig:canales} we observe that the investment levels at each channel vary largely in scale, with investments in \texttt{TV}, and \texttt{OOH} dominating the other channels.
\item The investment strategies adopted by the firm at each channel are also qualitatively different. Some of them show spikes while others depict a relatively even spread investment across time.
\end{itemize}
A handful of other predictors $X_t$ which are also known to affect sales will be used in the model, all of them weekly sampled:

\begin{itemize}
\item Global economic indicators: unemployment rate (\texttt{Unemp\_IX}), price index (\texttt{Price\_IX}) and consumer confidence index (\texttt{CC\_IX}).
\item Climate data: average weekly temperature (\texttt{AVG\_Temp}) and weekly rainfall (\texttt{AVG\_Rain}) along the country.
\item Special events: holidays (\texttt{Hols}) and important sporting events (\texttt{Sport\_EV}).
\end{itemize}

\subsection{Experimental setup}

Following the notation in Section \ref{sec:model}, we consider three model variants for the particular dataset in increasing order of complexity:


\begin{itemize}
\item \textbf{Baseline model}, which makes use of no external variables $$Y^{\text{B}}_t = Y_{NA,t} + Y_{T, t}.$$
\item \textbf{Auto regression}: this model  incorporates the external ambient and investment variables, so the equation of the model becomes 
$$ Y_t^{\text{RA}} = Y_{NA,t} + Y_{T, t} + Y_{R,t}.$$ We select an expected model size of 5 in the spike and slab prior, letting all variables to be treated equally.
\item \textbf{Regression (forcing)}: the model has same equation as before (we will refer to it as \text{RF}).
However, in the prior we force investment variables to be used by setting their corresponding $\pi_i = 1$, and imposing an expected model size of 5 for the rest of the variables.
\end{itemize}
In all cases, only the five principal advertising channels (\texttt{TV}, \texttt{OOH}, \texttt{ONLINE}, \texttt{SEARCH} and \texttt{RADIO}) will be used; the remaining two (\texttt{CINEMA} and \texttt{PRESS}) are sensibly lower both in magnitude and frequency than the others so we can safely disregard them in a first approximation.

As customary in a supervised learning setting with time series data, we perform the following split of our dataset: since it comprises four years of sales, we take the first two years of observations as training set, and the rest as holdout, in which we measure several predictive performance criteria. Before fitting the data, we scale the series to have zero mean and unit variance as this increases MCMC stability. Reported sales forecasts are transformed back to the original scale for easy interpretation. The models were implemented in R  using the \texttt{bsts} package \cite{scott2016bsts}.


\section{Discussion of results}

It is customary to aim at models achieving good predictive performance. 
For this reason, we test the predictive performance of our three models using two metrics:

\begin{itemize}
\item Mean Absolute Percentage Error: $$ \text{MAPE} = \frac{100\%}{T}\sum_{t=1}^T \frac{|y_t - \hat{y}_t|}{y_t},$$
where $y_t$ denotes the actual value; $\hat{y}_t$, the mean one-step-ahead prediction; and, $T$ is the length of the hold-out period.
\item Cumulative Predicted Sales over a year $Y$:
$$
\text{CPS}_{Y} = \sum_{t \in \mathcal{T}(Y)} \hat{y}_t
$$
where $\mathcal{T}(Y)$ denotes the set of time-steps $t$ contained in year $Y$.
\end{itemize}
These scores are reported in Table \ref{tab:mapes} with sales in million EUR. Note that, unsurprisingly, the models which include external information (RA and RF) achieve better accuracy than the baseline. In addition, we note that predictions are unbiased, since cumulative predictions are extremely close to their observed counterparts. Overall, we found the predictive performance of our models to be successful for a business scenario, as we achieve under 5\% relative absolute error using the variants augmented with external information. This is clearly useful for the decision maker who may forecast their weekly sales one week ahead to within a 5\% error in the estimation.

\begin{table}[H]
\centering
\begin{tabular}{|l|c|c|c|}
\hline
Model & MAPE & $\text{CPS}_{2013}$ & $\text{CPS}_{2014}$ \\
\hline
B &  5.85\% &   9660 & 9627 \\
RA &  4.62\% &  9680 & 9613 \\
RF &  4.59\% &  9665 & 9582 \\
\hline
\multicolumn{2}{|c|}{Cumulative True Sales:} & 9666 & 9610 \\
\hline
\end{tabular}
\caption{Accuracy measures for each model variant.} \label{tab:mapes}
\end{table}



\begin{figure}[h]
\centering
\includegraphics[scale=0.55]{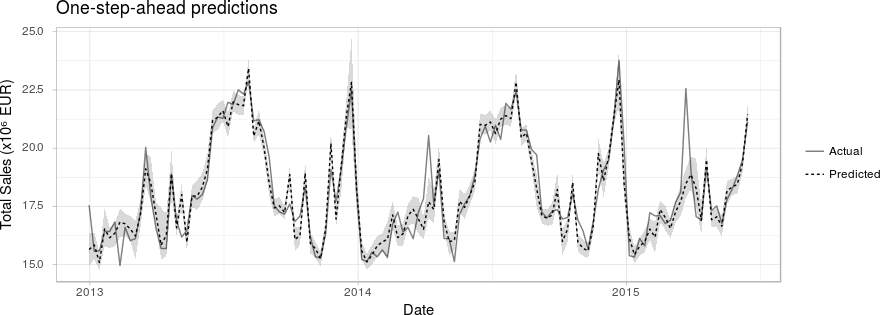}
\caption{One-step-ahead forecasts of the RF model versus actual data during hold-out period. 95\% predictive intervals are depicted in light gray.}\label{fig:forecasts}
\end{figure}

\begin{figure}[H]
\centering
\includegraphics[scale=0.6]{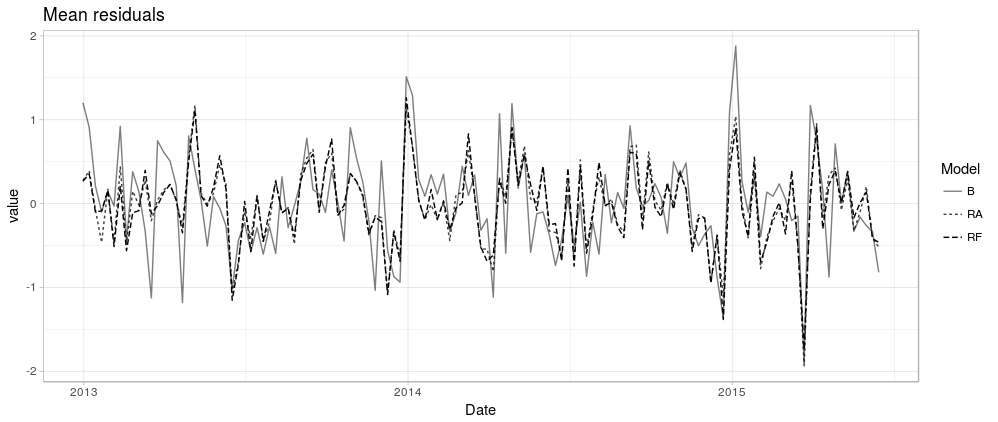}
\caption{Mean residuals for each model}\label{fig:residuals}
\end{figure}

Figure \ref{fig:forecasts} displays the predictive ability of model RF over the hold-out period. Note that the model is sufficiently flexible to adapt to fluctuations such as the peaks at Christmas. Predictive intervals also adjust their width with respect to the time to reflect varying uncertainty, yet in the worst cases they are sufficiently narrow. Further information can be tracked in Figure \ref{fig:residuals}, where mean standardized residuals are plotted for each model variant. Notice that the residuals for models RA and RF are roughly comparable, being both sensibly smaller than those of the baseline. This means that the simpler Nerlove-Arrow model benefits from the addition of the ambient variables $X_i$, as suggested in our findings from Table \ref{tab:mapes}.

Having built good predictive models, we inspect them more closely with the aim of performing valuable inferences for our business setting. The average estimated parameter and expected standard deviations of the $q_i$ coefficients for the different advertising channels are displayed in Table \ref{tab:1}. We also show the weights of the ambient variables $X_i$ for the augmented models RA and RF in Table \ref{tab:11}, as well as the probability of a variable being selected in the MCMC simulation for a given model in Figure \ref{fig:inc_probas}. Convergence diagnostics of the MCMC scheme are reported in Appendix \ref{app:GR}.


\begin{table}[h]
\centering
\begin{tabular}{ |l|c|c|c|c|c|c|c| }
  \hline
  & \multicolumn{2}{|c|}{Model B} & \multicolumn{2}{|c|}{Model RA} & \multicolumn{2}{|c|}{Model RF}\\
  \hline
  Channel & mean & sd & mean & sd & mean & sd\\
  \hline
  \texttt{y\_AR} & \textbf{8.00e-01} & \textbf{5.10e-02} &  \textbf{5.17e-01} &  \textbf{4.50e-02} &  \textbf{5.07e-01} &  \textbf{4.55e-02}   \\
  \texttt{OOH} & 9.30e-03 & 3.11e-02 & 1.15e-03 & 9.54e-03  & \textbf{6.10e-02} &  \textbf{3.54e-02}\\
  \texttt{ONLINE} & 1.10e-03 & 9.25e-03  & 1.95e-03 &  1.23e-02 & \textbf{9.04e-02} & \textbf{4.01e-02}\\
  \texttt{RADIO} & -5.34e-04 & 6.80e-03  &-2.61e-05 & 2.30e-03 &   -2.57e-02 & 3.42e-02\\
  \texttt{TV} & -2.85e-04 & 5.12e-03 & -5.53e-05 & 3.71e-03  & -6.14e-02 & 4.21e-02\\
  \texttt{SEARCH} & 1.51e-05  & 2.74e-03  & 8.58e-05 & 2.49e-03  & 1.38e-02 & 3.50e-02 \\
    \hline
\end{tabular} \caption{Expected value and standard error of coefficients $q_i$. Statistically significant coefficients appear in bold.}\label{tab:1}
\end{table}

\begin{table}[H]
\centering
\begin{tabular}{ |l|c|c|c|c|c| }
  \hline
  &  \multicolumn{2}{|c|}{Model RA} & \multicolumn{2}{|c|}{Model RF}\\
  \hline
  Channel & mean & sd & mean & sd\\
  \hline
  \texttt{Sport\_EV} & \textbf{-2.06e-01} & \textbf{3.80e-02} & \textbf{-2.00e-01} &\textbf{ 3.71e-02} \\
  \texttt{AVG\_Temp} &  \textbf{2.57e-01} & \textbf{4.43e-02} & \textbf{2.43e-01} & \textbf{4.65e-02}   \\
  \texttt{Hols} & \textbf{3.10e-01} &\textbf{ 3.70e-02}   & \textbf{3.15e-01} & \textbf{3.72e-02}  \\
  \texttt{AVG\_Rain} &  -2.86e-02 & 5.01e-02 & -1.96e-02 & 4.14e-02     \\
  \texttt{Price\_IX} &  1.38e-03 &  1.15e-02 & 3.60e-03 & 1.92e-02  \\
  \texttt{Unemp\_IX} &  6.34e-05 & 3.32e-03 & 2.36e-04 & 7.79e-03  \\
  \texttt{CC\_IX} & 6.27e-07 & 2.94e-03 & 2.85e-04 & 5.50e-03 \\
    \hline
\end{tabular} \caption{Expected value and standard error of coefficients $\beta_i$. Statistically significant coefficients are in bold.}\label{tab:11}
\end{table}


\begin{figure}[H]
\centering
\includegraphics[scale=0.6]{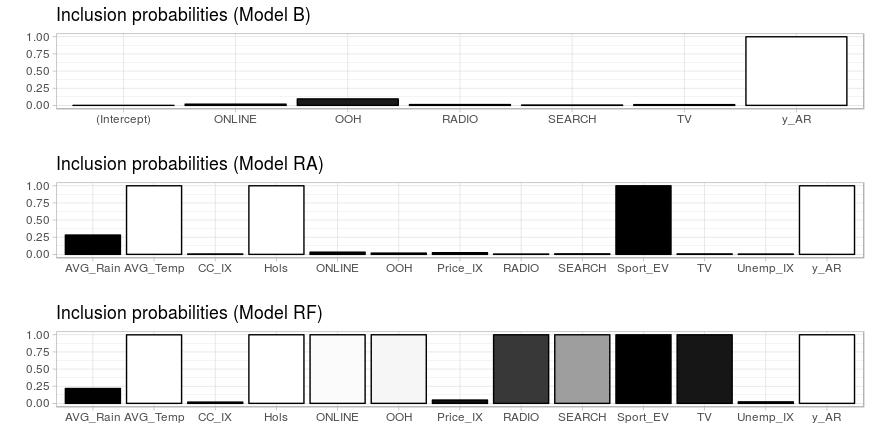}
\caption{Selection probabilities in the MCMC simulation for each predictor variable in the three models. The color code shows variables positively (white) or negatively (black) correlated with sales. }\label{fig:inc_probas}
\end{figure}


Looking at the ambient variables, the following comments seem in order:
\begin{itemize}

\item From Table \ref{tab:11} we see that the socio-economic indicators (unemployment rate, inflation and consumer confidence) do not seem statistically relevant for the problem at hand. 

\item Looking at the sign of the coefficients of the most significant regressors $X_i$ (\texttt{Hols}, \texttt{Sport\_EV}, \texttt{AVG\_Temp} and \texttt{AVG\_Rain}) we see that they are as we would naturally expect. Moreover, their absolute value is well above the error in both models RA and RF, a strong indicator of their influence in the expected weekly sales (\emph{cf.} Table \ref{tab:11}, Figure \ref{fig:inc_probas}). 

\item We see, for instance, that sporting events are negatively correlated with sales. This can be interpreted as follows: major sporting events in the country where the data have been recorded receive a large media coverage and are followed by a significant fraction of the population. The restaurant chain in this study has no TVs broadcasting in their restaurants, so customers probably choose alternative places to spend their time on a day when \texttt{Sport\_EV} = 1. 

\item The sign of \texttt{Hols} and $\texttt{AVG\_Temp}$ is positive, showing strong evidence for the fact that sales increase in periods of the year where potential customers have more leisure time, like national holidays or the summer vacation. 

\item One would expect \texttt{AVG\_Rain} to be negatively correlated with restaurant sales, but in our study (despite having negative sign) it is not statistically significant. A possible explanation is that \texttt{AVG\_Rain} records average rainfall over a large country. A more fine-grained analysis on a restaurant by restaurant basis incorporating local weather conditions will be performed in a future study, probably revealing a larger influence of this variable on the sales prediction.

\end{itemize}

Next, we turn our attention to the investment variables across different advertising channels.

\begin{figure}[h]
\centering
\includegraphics[scale=0.75]{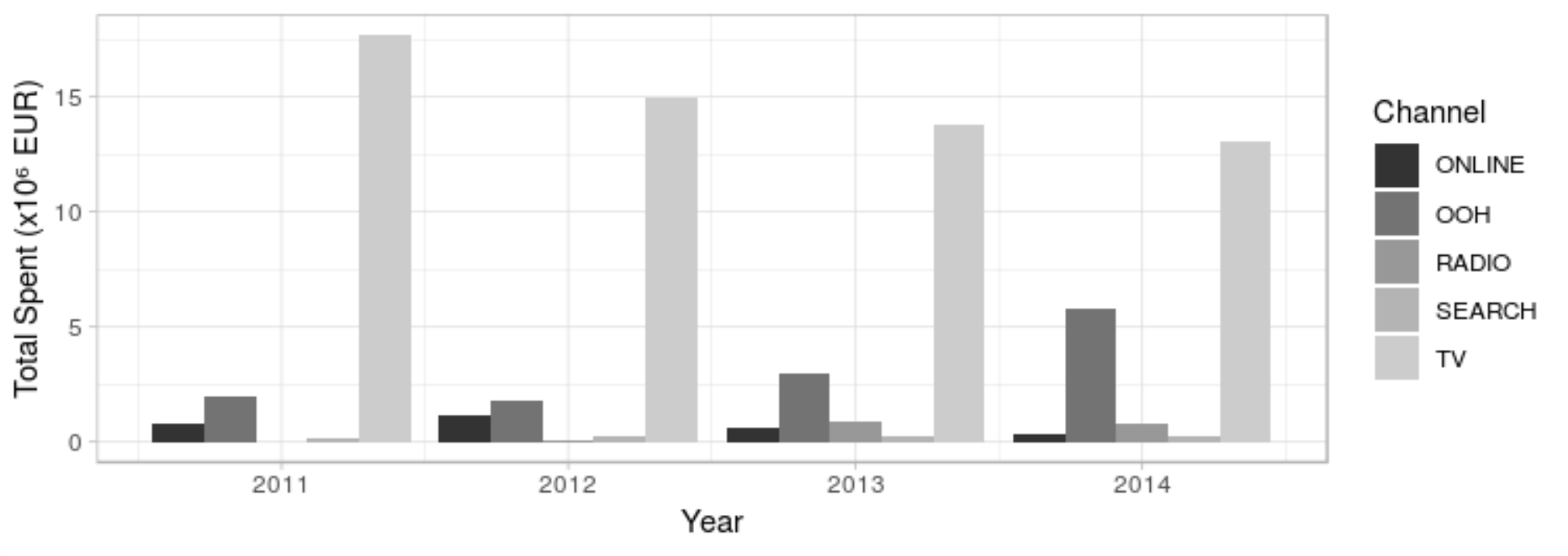}
\caption{Yearly total expenditures on advertising campaigns for the restaurant chain, per advertising channel.}\label{fig:yearly}
\end{figure}

\begin{itemize}

\item The advertising channels $u_i$ are almost never selected in model RA, and their $q_i$ coefficients are not significantly higher than their errors to be considered influential in the model. In model RF, however, there is strong evidence that their effect is more than a random fluctuation (\emph{cf.} Table \ref{tab:1}, Figure \ref{fig:inc_probas}). 
\item The negative sign in both \texttt{RADIO} and \texttt{TV} in all three models suggests that (at least locally) part of the expenditures in these two channels should be diverted towards other channels with positive sign on their $q_i$ coefficients, specially to the channel with the strongest positive coefficient (\texttt{ONLINE} and \texttt{OOH}).
\item It is interesting that the trend that shows the year-to-year advertising budget of this firm (\emph{cf.} figure \ref{fig:yearly}) has a significant reduction in \texttt{TV} expenditures and a big increase in \texttt{OOH}. \texttt{RADIO} however is not reduced accordingly but increased, and \texttt{ONLINE} ---\,which our model considers the best local inversion alternative\,--- follows the inverse path. It has to be noted, however, that \texttt{ONLINE} expenditures are typically correlated with discounts, coupons and offers, and this information was not available to us.
\item The autoregressive term is close to $0.5$, which means that the immediate effect of advertising is roughly half to the \textit{long run accumulated effects}.
\end{itemize}

\subsection{Budget allocation. Model-based solutions}\label{section: budget allocation}

We propose a model which can be used as a decision support system for the manager, helping her in adopting the  investment strategy in advertising. The company is interested in maximizing the expected sales for the next period, subject to a budget constraint for the advertising channels and also a risk constraint, i.e., the variance of the predicted sales must be under a certain threshold.
This optimization problem based on one-step-ahead forecasts can be formulated as a non linear, but convex, problem that depends on the parameter $\sigma^2$:

\begin{equation*}
\begin{aligned}
& \underset{u_{(t+1),1}...u_{(t+1),k}}{\text{maximize}}
& &  E[ \bar{y}_{t+1} | y_{1:t}, x_{t+1}, u_{t+1}] \\
& \text{subject to}
& & \sum_{i=1}^k  u_{(t+1),i} \leq b_{t+1} \\
& & & Var[ \bar{y}_{t+1} | y_{1:t}, x_{t+1}, u_{t+1}] \leq \sigma^2, \\
\end{aligned}
\end{equation*}
where $b_t$ is the total advertising budget for week $t$ and $\sigma$ is a parameter that controls the risk of the sales. We have made explicit the dependence on the regressor variables $x_t$ and advertisement investments $u_t$ in the mean and variance expressions.
Solving for different values of $\sigma$, we obtain a continuum of Pareto optimal investment strategies that we can present to the manager, each one representing a different trade-off between risk and expected sales that we can plot in a risk-return diagram.
This approach greatly reduces the decision space for the manager.

A possible alternative would be to rewrite the objective function as 
$$E[ \bar{y}_{t+1} | y_{1:t}, x_{t+1}, u_{t+1}] - \lambda \sqrt{Var[ \bar{y}_{t+1} | y_{1:t}, x_{t+1}, u_{t+1}]}$$ 
which may be regarded as a lower quantile if the predictive samples $\bar{y}^{(i)}_{t+1}$ are normally distributed.
As in the previous approach, different values of $\lambda$ represent different risk-return trade-offs.

If the errors in the observation equation (\ref{eq:obs}) are Normal, the computations for expected predicted sales and the above variance can be done exactly and quickly using conjugacy as in \cite[3.7.1]{zbMATH06123712}.
Otherwise, the desired quantities can be computed through Monte Carlo simulation, as in Section \ref{s:MCMC}.

Note that due to the nature of the state-space model, it is straightforward to extend the previous optimization problem over $k$ timesteps, for $k>1$.
The two objectives would be the expected total sales and the total risk over the period $t+1,\dots,t+k$.
For normally distributed errors, the predicted sales on the period also follow a Normal distribution, which makes computations specially simple.

The above optimization problem may not need to be solved by exhaustively searching over the space of possible channel investments. In a typical business setting, the manager would consider a discrete set of $S$ investment strategies that are easy to interpret, so she may perform $S$ simulations of the predictive distribution and use the above strategy to discard the Pareto suboptimal strategies.

\section{Summary and future work}

We have developed a data-driven approach for the management of advertising investments of a firm. First, using the firm's investment levels in advertising, we propose a formulation of the Nerlove-Arrow model via a Bayesian structural time series to predict an economic variable (global sales)  which also incorporates information from the external environment (climate, economical situation and special events). The model thus defined offers low predictive errors while maintaining interpretability and can be built in a modular fashion, which offers great flexibility to adapt it to other business scenarios. The model performs variable selection and allows to incorporate \emph{prior} information via the \emph{spike-and-slab} prior. It can handle non-gaussian deviations and also provides hints to which of the advertising channels are having positive effects upon sales. The model can be used as a basis for a decision support system by the manager of the firm, helping with the task of allocating ad investments.

Possible extensions of this model include:

\begin{enumerate}
\item Use a different model to explain the influence of advertising upon sales. This model could take into account interactions between the channels or allow for different long-term effects for each of them.
\item Develop a model-based strategy for long-run temporal and cross-sectional budget allocation.
\item Model each restaurant individually instead of using total aggregated sales. For this approach to be sound, it would entail to use both  local values of the ambient variables $X_i$ and some of the investment  channels $u_i$ (e.g. \texttt{OOH}).
\item Include in the model the effect of special discounts, promotions and coupons, since these are probably highly correlated with some of the channels, specially \texttt{ONLINE} and \texttt{SEARCH}.
\end{enumerate}

\section*{Acknowledgements}

The authors acknowledge financial support from the Spanish Ministry of Economy and Competitiveness, through the ``Severo Ochoa Programme for Centres of Excellence in R\&D'' (SEV-2015-0554).  V.G. acknowledges support for grant FPU16/05034. The research of D.G-U is supported in part by Spanish MINECO-FEDER Grant  MTM2015-65888-C4-3 and MTM2015-72907-EXP. 
The authors also thank the members of the SPOR-Datalab group at ICMAT for their suggestions and support.

\bibliography{00_DLMs_Biblio}
\bibliographystyle{unsrt}

\newpage
\appendix

\section{MCMC Convergence Diagnostics} \label{app:GR}

\begin{figure}[h]
\centering
\includegraphics[scale=0.6]{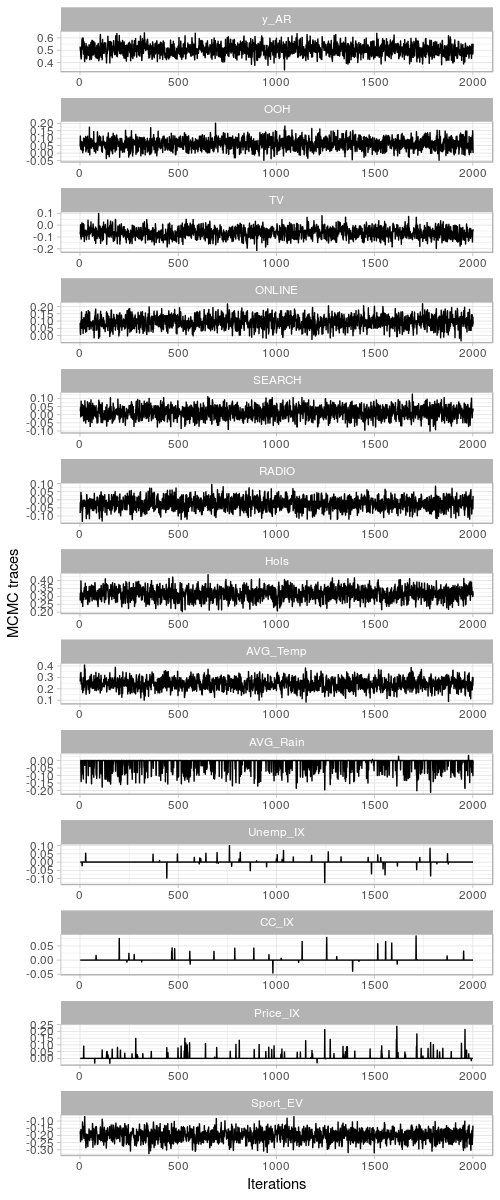}
\caption{MCMC traces after a burn-in of 2000 iterations.}\label{fig:traces}
\end{figure}

In order to asses the convergence of the MCMC scheme described in Section \ref{s:MCMC}, we used the Gelman-Rubin convergence statistic $\hat{R}$. We report its value for each latent dimension of our best performing model, the RF variant, in Table \ref{tab:GR}. All values are under 1.1, confirming correct convergence. In addition, we display the trace plots for each variable at Figure \ref{fig:traces}.

\begin{table}[H]
\centering
\begin{tabular}{|l|c|c|c|c|c|c|c|}
\hline
coefficient & \texttt{y\_AR} & \texttt{OOH} & \texttt{TV} & \texttt{ONLINE} & \texttt{SEARCH} & \texttt{RADIO} & \texttt{Hols} \\
\hline 
$\hat{R}$ & 0.9997814 & 0.9997621 & 1.00001 & 0.9998523 & 1.000033 & 1.000522 & 0.9999129 \\
\hline
coefficient & \texttt{AVG\_Temp} & \texttt{AVG\_Rain} & \texttt{Unemp\_Ix} & \texttt{CC\_IX} &  \texttt{Price\_IX}& \texttt{Sport\_EV}  & \\
\hline
$\hat{R}$ & 0.999768 & 0.9997925 & 1.030499 & 1.039316 & 1.005669 & 1.000532 & \\
\hline
\end{tabular}\caption{Gelman-Rubin statistic results}\label{tab:GR}
\end{table}

\end{document}